\documentclass[letterpaper, 10 pt, journal, twoside]{IEEEtran}

\IEEEoverridecommandlockouts                              

\usepackage{graphicx}
\usepackage{amsmath,amssymb,amsfonts}
\usepackage{algpseudocode,algorithm}
\usepackage{subcaption}
\usepackage{makecell}
\usepackage{cite}
\usepackage{url}
\usepackage{multirow}
\usepackage{tabularx}

\begin{document}

\title{
\LARGE \bf Redesigning Regularization for Effective Policy Smoothing
}

\author{Taisuke Kobayashi$^{1}$ and Naoto Yamanaka$^{1}$
\thanks{\copyright 2026 IEEE.  Personal use of this material is permitted.  Permission from IEEE must be obtained for all other uses, in any current or future media, including reprinting/republishing this material for advertising or promotional purposes, creating new collective works, for resale or redistribution to servers or lists, or reuse of any copyrighted component of this work in other works.}
\thanks{Manuscript received: June, 11, 2026; Revised August, 27, 2026; Accepted October, 8, 2026.}
\thanks{This paper was recommended for publication by Editor Sehoon Ha upon evaluation of the Associate Editor and Reviewers' comments.
This work was supported by JST PRESTO, Japan, Grant Number JPMJPR2514 and JSPS KAKENHI, Grant-in-Aid for Scientific Research(S), Grant Number JP22H05002.}
\thanks{$^{1}$T. Kobayashi and N. Yamanaka are with the National Institute of Informatics (NII) and with The Graduate University for Advanced Studies (SOKENDAI),
        2-1-2 Hitotsubashi, Chiyoda-ku, Tokyo, 101-8430, Japan
        {\tt\small \{kobayashi, naoto\_yamanaka\}@nii.ac.jp}}%
\thanks{Digital Object Identifier (DOI): see top of this page.}
}

\markboth{IEEE Robotics and Automation Letters. Preprint Version. Accepted October, 2026}
{Kobayashi and Yamanaka: Redesigning Regularization for Effective Policy Smoothing}

\maketitle

\begin{abstract}

This paper proposes a novel regularization design to effectively smooth policy functions in reinforcement learning.
While regularization that enhances ``global'' Lipschitz continuity was initially considered, it has been limited to ``local'' Lipschitz continuity due to a tradeoff between smoothness and expressiveness.
However, it has become apparent that the original implementation is cumbersome and does not provide sufficient smoothing, leading to a preference for simpler implementations.
This stems from a discrepancy between theory and implementation, and a more appropriate implementation can expect to facilitate smoothing.
Therefore, this paper identifies three reasons why the original implementation does not function adequately and provides remedies for them.
This modified regularization performs well across multiple tasks and algorithms, successfully achieving smooth motion with sufficient control performance.
Furthermore, by applying it to sim-to-real reinforcement learning for a quadruped robot, it is demonstrated that smooth and energy-efficient motion provides robustness against sudden changes in target velocity commands.

\end{abstract}


\section{Introduction}

\IEEEPARstart{R}{einforcement} learning (RL)~\cite{sutton2018reinforcement} is widely used as a fundamental technology underpinning modern robot control.
By appropriately designing a reward function that measures task performance, robots can learn control policies that cannot be derived mathematically, thereby enabling them to perform complex movements.
In particular, by acquiring vast amounts of experience in diverse simulation environments, it has become possible to learn efficiently while minimizing the cost of operating robots in the real world~\cite{rudin2022learning,hu2023simulation,radosavovic2024real}.

However, the policy deployed on real robots often causes jitter, failing stable control.
This problem stems from the fact that RL policies are overly sensitive to changes in the state inputs~\cite{mysore2021regularizing}.
Even slight observation noise or deviations from expected behavior can trigger extreme actions, leading to oscillations and divergence.

Consequently, methods for smoothing policies (or the value functions that generate their learning signals) have been widely studied.
An important concept that characterizes this smoothing is Lipschitz continuity, and the degree of smoothness is quantified by the smallness of an upper bound known as the Lipschitz constant.
To minimize the Lipschitz constant, two approaches have been explored: i) imposing constraints and optimizing the policy during the design phase of the policy’s functional approximation~\cite{gouk2021regularisation}; and ii) applying regularization during training~\cite{mysore2021regularizing} (details are provided in the next section).
Since the former approach increases computational costs during inference and may adversely affect the robot's realtime control, this paper focuses on regularization.
Note that although ad-hoc implementations that add terms related to the smoothness to the reward function are widely adopted~\cite{mysore2021train,siekmann2021sim}, we exclude them from our discussion since it is well known that they are highly dependent on the target task (i.e. the other reward terms) and rarely yield sufficient results~\cite{mysore2021regularizing}.

One of the challenges in the regularization methods is the tradeoff between smoothness and expressiveness.
Specifically, applying regularization that smooths policies (and actions) across the whole state space results in a loss of expressiveness~\cite{mysore2021regularizing}.
The design-based methods avoided this tradeoff by automatically optimizing the degree of smoothness based on state inputs~\cite{song2023lipsnet}, which corresponds to the consideration of ``local'' Lipschitz continuity.
This concept also appears in the regularization methods, and locally Lipschitz continuous constraint (L2C2)~\cite{kobayashi2022l2c2} introduced it with the local space based on the start and end states of a single transition.
Since this transition is commonly included in RL's experience data, L2C2 is applicable to most of RL algorithms.

However, L2C2 sometimes could not provide sufficient smoothing~\cite{kwak2026enhancing} (this is also observed from our experiments).
Perhaps because the benefits are minimal for the cumbersome implementation, a simplified version has been adopted in several applied research studies~\cite{huang2025learning,huang2026towards,zhao2026agile,yu2026mastering}.
This finding suggests that the issue lies not so much with the theory behind L2C2, but rather with the original method used to implement it.
Although alternative regularization methods to L2C2 have also been designed (see the next section), they often require modifications in the algorithms applied beyond simply adding a regularization term~\cite{lee2024gradient,kwak2026enhancing}.
Therefore, prioritizing generality, this paper does not consider them.

In this paper, we reveal the following three implementation issues in L2C2, followed by their corresponding solutions:
\begin{enumerate}
    \item Design of distance function in state space:
    The original $L_\infty$ norm tended to be nearly one in high-dimensional state spaces~\cite{david2004order}, failing the implicit weighting.
    Therefore, we adjust the distance function so that its expected value can approximately be specified.
    \item Selecting divergence between policies:
    The original Hellinger distance vanished gradients around its upper bound, disabling the regularization.
    Therefore, we switch to Kullback-Leibler (KL) divergence approximated by a non-negative Monte Carlo method~\cite{nielsen2020non}.
    \item Scalarization of batched loss:
    The original mean operation conflicted with Lipschitz continuity that caps the upper bound, being unable to sufficiently regularize some worst-case data.
    Therefore, we introduce a tilted loss that prioritizes worst-case data~\cite{li2023tilted,lin2024smooth}.
\end{enumerate}
The new L2C2 implementation incorporating these modifications (so-called L2C2-v2) achieves stronger smoothing while maintaining control performance across three benchmark tasks with two RL algorithms.
Furthermore, when L2C2-v2 was applied to the locomotion control of a quadruped robot using sim-to-real RL, it successfully produced a smoother and energy-efficient walking while also providing robustness against sudden changes in target velocity commands.

This paper makes the following three contributions:
\begin{enumerate}
    \item We implemented a new L2C2 that enables effective smoothing from a theoretical perspective.
    \item We exemplified its versatility in improving the control performance with appropriate smoothing across multiple benchmark tasks with multiple RL algorithms.
    \item We demonstrated that the proposed method makes the real-world robot's policy smooth, energy-efficient, and robust against sudden changes in conditions.
\end{enumerate}

\subsection{Related work}

The policy smoothing discussed in this paper can be broadly classified into two approaches (or combinations thereof~\cite{christmann2024benchmarking,xie2026learning}), which are summarized below.
It should be noted that, in addition to these, there have been several studies that indirectly smooth policies by smoothing the value function used to compute policy learning signals~\cite{kobayashi2022l2c2,harder2024continuity,lee2026stabilizing}.
These are considered complementary approaches and are therefore omitted from this paper.

\subsubsection{Modeling smoothness-aware policy}

Since function approximation using naive neural networks has a risk of abrupt changes in outputs~\cite{gouk2021regularisation}, models capable of generating smoother outputs have been developed.
The most representative example is LipsNet~\cite{song2023lipsnet}, which uses an automatic differentiation to calculate the gradient of the outputs at once.
With it and a learnable coefficient, the outputs are rescaled to achieve the desired gradient scale.
Regularizing the coefficient during training makes the policy smooth to the extent that it does not hinder the completion of tasks~\cite{zhang2025robust}.
Furthermore, by making the coefficients dependent on the state inputs, it is easy to transform the model from global Lipschitz continuity to local Lipschitz continuity.
Other models that have been designed include unique network models that apply filtering to internal features~\cite{wang2024smooth,wang2025ode}, as well as models that decompose and resynthesize state inputs in the frequency domain~\cite{song2025lipsnet++}.

While these models are highly practical in that they can be expected to provide a certain degree of smoothness at the model level, they inevitably incur higher computational costs compared to naive models, making them ill-suited for realtime robot control (particularly when using embedded systems with limited computational resources).
Therefore, this paper omits this approach from further discussion.

\subsubsection{Applying smoothness regularization}

Numerous studies have designed their own regularization terms, which are added into the policy loss, to promote smoothness explicitly.
Initially, spatio-temporal regularization terms were designed~\cite{mysore2021regularizing}; however, these imposed a form of global Lipschitz continuity, compromising the necessary expressiveness.
Subsequent research, therefore, focused on designing regularization based on local Lipschitz continuity.
In particular, L2C2, the baseline method discussed in this paper, has established a basis for this local space based on start and end points of transitions~\cite{kobayashi2022l2c2}.
This allows the system to require the same behavior during transitions, while permitting arbitrary behavior outside those intervals.
However, because the original implementation was cumbersome yet ineffective, simplified implementations have been adopted for applications such as humanoid control~\cite{huang2025learning,huang2026towards,zhao2026agile,yu2026mastering}.
Therefore, in this paper, we evaluate both the original implementation and a simplified version as baselines.

In addition, methods have been proposed to ensure a certain level of expressiveness by considering global Lipschitz continuity with respect to the magnitude of action changes (using numerical differentiation~\cite{lee2024gradient} or automatic differentiation~\cite{chen2025learning}).
Alternative regularization has been devised to minimize the error between the actual action and the expected action at the next time step, thereby ensuring that subsequent actions do not diverge too widely~\cite{kwak2026enhancing}.
These approaches require extensions to the experience data structure~\cite{lee2024gradient} or the addition of a prediction head~\cite{kwak2026enhancing}, necessitating modifications beyond the introduction of a regularization term and thus lacking in generality.
Only Lipschitz-constrained policies (LCP)~\cite{chen2025learning}, with the gradient norm of the policy log-likelihood as a regularization term, can be applied to any RL algorithms, in exchange for the cost of second-order derivatives.
Therefore, this paper adds LCP to one of the baselines.

Thus, since general-purpose smoothing techniques are still limited, revisiting and improving L2C2 is practical and worthy of research.
Therefore, this paper focuses its discussion on policy-smoothing regularization on the local space and without altering the structure of the empirical data.

\section{Preliminaries}

\subsection{Reinforcement learning}

RL~\cite{sutton2018reinforcement} aims to optimize an agent to reach a desired situation by repeatedly interacting with an unknown environment.
This interaction (or transition) is given as Markov decision process (MDP) with state $s \in \mathcal{S}$, action $a \in \mathcal{A}$, and reward $r \in \mathbb{R}$.
The action $a$ is determined according to a trainable probability distribution (named policy), $\pi(a \mid s)$.
According to $a$, $s$ is shifted to the next one, $s^\prime$, under the state-transition probability.
Then, one transition is evaluated by $r$.
Note that the experience data in RL is usually structured by $(s, a, s^\prime, r)$.

With this framework, the sum of rewards into the future (named return), $R_t = \sum_{k=0}^\infty \gamma^k r_{t+k}$ with $\gamma \in [0, 1)$ the discount factor, is the target of RL optimization.
Specifically, the expected $R$ is well-known as the value function, $V(s) = \mathbb{E}[R_t \mid s_t=s]$ or $Q(s,a) = \mathbb{E}[R_t \mid s_t=s, a_t=a]$, and $\pi$ is optimized to maximize them.
Although this optimization method branches into several algorithms, this paper focuses on actor-critic algorithms that explicitly model $\pi$ and can handle continuous action spaces, which are suitable for robot control.
In this case, the mean of policy with respect to the input, $\mu_\pi(s)$, is often utilized to deterministically generate continuous actions.
Here, let us define the loss function to be minimized for optimizing $\pi$ as $\mathcal{L}_\pi$.

\subsection{Locally Lipschitz continuous constraint: L2C2}

Since the optimization of $\pi$ is performed based on the estimated value function, $V(s)$ and/or $Q(s,a)$, it is prone to producing extreme outputs due to estimation errors.
In the real world, where observation noise is present, this manifests as jitter, which degrades the robot's motion performance (and breaks the robot or environment in the worst case).
L2C2 aims to mitigate that by the policy smoothing~\cite{kobayashi2022l2c2}.

Specifically, L2C2 considers the local Lipschitz continuity.
\begin{align}
    \cfrac{d_{Y}(f(x_1), f(x_2))}{d_{X_i}(x_1, x_2)} = \rho_f(x_1, x_2) \leq K_i
    \label{eq:def_llc}
\end{align}
where, $x_{1,2} \in X_i$ denote the inputs in the $i$th local space $X_i$, which map to $Y$ by $f$.
These two spaces have their own distance functions, $d_{X_i, Y}$.
If $K_i$, so-called the Lipschitz constant, exists for all pairs of $x_1$ and $x_2$, the local Lipschitz continuity holds.
In other words, the maximum ratio, $\max_{x_{1,2} \in X_i} \rho_f(x_1, x_2)$, must be smaller than or equivalent to $K_i$.
Note that for the policy smoothing, $x$ corresponds to $s$ and $f$ is $\pi$, while $X_i$, $d_{X_i}$, $d_{Y}$ need to be designed additionally (see later).

If we know the desired $K_i$ for all local spaces, the above condition can be given as inequality constraint for the optimization problem of RL with $\mathcal{L}_\pi$.
However, it is infeasible since numerous local spaces, all of which require different $K_i$, exist.
L2C2 therefore converts the constraint to the corresponding regularization as follows:
\begin{align}
    \begin{split}
        \pi^\ast(\cdot \mid s) &= \arg\min_{\pi} \mathcal{L}_\pi(s) + \lambda \max_{\tilde{s} \sim \mathcal{S}_{s,s^\prime}}\rho_\pi(s, \tilde{s})
        \\
        \mathrm{s.t.} \ \mathcal{S}_{s,s^\prime} &= \{ \tilde{s} \mid d_{\mathcal{S}_{s,s^\prime}}(s, \tilde{s} \mid \sigma \Delta s) \leq 1 + \nu \}
    \end{split}
    \label{eq:l2c2_ideal}
\end{align}
where, $\lambda \geq 0$ denotes the gain, $d_{\mathcal{S}_{s,s^\prime}}: \mathcal{S}_{s,s^\prime} \times \mathcal{S}_{s,s^\prime} \mapsto [0, 1+\nu]$ denotes the distance of local state space with $\sigma > 0$ is for the scaling factor (normally, $\sigma=1$), and $\Delta s = s^\prime - s$.
In addition, $\nu > 0$ is added for numerical stability (see later).

The practical sampling method of $\tilde{s}$ is given below.
\begin{align}
    \begin{split}
        \tilde{s} &= s + \sigma \Delta s \epsilon
        \\
        \epsilon &\sim \mathcal{U}_{|\mathcal{S}|}(-1, 1) \ \mathrm{s.t.} \ \epsilon \in \mathcal{S}_{0,1}
    \end{split}
    \label{eq:sample_sl}
\end{align}
Note that the original implementation uses $|\mathcal{S}|$-dimensional uniform distribution $\mathcal{U}_{|\mathcal{S}|}$, while the simplified version modifies it to univariate uniform distribution.

Anyway, eq.~\eqref{eq:l2c2_ideal} is still intractable to solve, as finding the maximum ratio is computationally expensive.
Therefore, it is approximated using batched data to compute $\mathcal{L}_\pi$ as follows:
\begin{align}
    \theta^\ast = \arg\min_{\theta} & \mathbb{E}_{(s,a,s^\prime,r) \sim D}[\ell_\pi(s,a,s^\prime,r)]
    \nonumber \\
    & + \mathbb{E}_{(s,s^\prime) \sim D; \tilde{s} \sim \mathcal{S}_{s,s^\prime}}[\lambda \rho_\pi(s, \tilde{s})]
\end{align}
where, $\theta$ denotes the parameters to model $\pi$, $D$ is the replay buffer, and $\ell_\pi$ is the loss for each experience.

In addition, to compute $\rho_\pi$, the original L2C2 sets $d_{\mathcal{S}_{s,s^\prime}}$ as $L_\infty$ norm extended by $\nu$, $\|\epsilon\|_\infty + \nu$, and $d_{\pi}$ as Hellinger distance (more specifically, its one-sample Monte Carlo approximation).
With this setting, the sampling of $\epsilon$ from $\mathcal{U}_{|\mathcal{S}|}(-1,1)$ in eq.~\eqref{eq:sample_sl} always satisfies the condition, which can be ignored now.
In addition, the following heuristic is introduced to stabilize the computation and to intuitively design hyperparameters.
\begin{align}
    w(\epsilon)=\cfrac{\lambda}{\|\epsilon\|_\infty + \nu} \in [\underline{\lambda}, \overline{\lambda}]
    \Rightarrow  \nu = \cfrac{\underline{\lambda}}{\overline{\lambda} - \underline{\lambda}}, \ \lambda = \overline{\lambda} \nu
    \label{eq:design_lambda}
\end{align}
In other words, since the denominator of $\rho_\pi$ corresponds to the weight, $\lambda \rho_\pi$ is separated to the optimization target and its weight.
Then, $\lambda$ and $\nu$ are designed so that the weight falls within the specified upper and lower bounds $[\underline{\lambda}, \overline{\lambda}]$.
With this design, one can expect that the closer $\tilde{s}$ is to $s$, the stronger the smoothing effect is while preventing the weight from diverging; conversely, the further $\tilde{s}$ is from $s$, the weaker the smoothing effect is, while not losing the weight.

\section{Proposal}

\subsection{Issues in the original implementation}

As mentioned earlier regarding L2C2, it was confirmed that the expected smoothing did not occur~\cite{kwak2026enhancing}, which is also demonstrated in the experiments described later.
We assume that the following three implementation issues remain.
\begin{enumerate}
    \item Order statistics~\cite{david2004order} explains that $L_\infty$ norm for $\epsilon$ sampled from $|\mathcal{S}|$-dimensional uniform distribution is equal to the value sampled from a beta distribution, $\mathcal{B}(|\mathcal{S}|, 1)$.
    Consequently, when the state space is large, most of the weights become near $\underline{\lambda}$ with $\|\epsilon\|_\infty \simeq 1$, thereby weakening the smoothing effect.
    \item Hellinger distance, which L2C2 adopted as a metric to measure the divergence between policies, is known to have a range of $[0, 1]$.
    Although a one-sample Monte Carlo approximation can exceed this upper bound, this is not the case from a statistical perspective.
    Consequently, when there is a significant divergence between the policies for $s$ and $\tilde{s}$, the regularization term saturates and the gradient vanishes, thereby preventing the smoothing.
    \item Although L2C2 approximates $\max_{\tilde{s}} \rho_\pi(s, \tilde{s})$, which we wish to consider under Lipschitz continuity, by $\mathbb{E}_{\tilde{s}}[\rho_\pi(s, \tilde{s})]$, this approach attempts to make all data points smooth equally.
    Consequently, due to the bias-variance tradeoff~\cite{geman1992neural}, it is unable to sufficiently suppress the worst-case data that should be smoothed with higher priority (and continues to smooth even points that are already sufficiently smooth).
\end{enumerate}
Hence, the original L2C2 was ineffective because it failed to perform smoothing when it was most needed.
In this paper, we propose solutions to these issues one-by-one.

\subsection{Local state-space distance}

First, we address the issue that, in high-dimensional state spaces, most samples result in $\|\epsilon\|_\infty \simeq 1$, causing the weights to become nearly fixed at $\underline{\lambda}$.
Specifically, we redesign the distance function on the local state space, $d_{\mathcal{S}_{s,s^\prime}}$.
In doing so, the practical designs in the original L2C2, namely, the simple $\epsilon$ sampling without validity check and the stable design of hyperparameters with $[\underline{\lambda}, \overline{\lambda}]$, are carried over.

Therefore, we modify $d_{\mathcal{S}_{s,s^\prime}}$ as follows:
\begin{align}
    d_{\mathcal{S}_{s,s^\prime}} = \|\epsilon\|_\infty^\eta + \nu
\end{align}
where, $\eta > 0$ is a design parameter.
Since this change does not alter the domain of the distance, eq.~\eqref{eq:design_lambda} still holds.

The added $\eta$ is designed such that the expected value of the weight computed with the above distance approximately matches the specified value (for simplicity, $\sqrt{\underline{\lambda}\overline{\lambda}}$).
Using the fact that $\|\epsilon\|_\infty = d_\infty \sim \mathcal{B}(|\mathcal{S}|, 1)$ and that $\nu \ll 1$, it can be approximately derived.
\begin{align}
    \sqrt{\underline{\lambda}\overline{\lambda}} &= \mathbb{E}_{d_\infty \sim \mathcal{B}(|\mathcal{S}|, 1)}\left[ \cfrac{\lambda}{d_\infty^\eta + \nu} \right]
    \nonumber \\
    &\simeq \int_0^1 d_\infty^{|\mathcal{S}| - \eta - 1} \lambda |\mathcal{S}| d d_\infty
    = \cfrac{\lambda |\mathcal{S}|}{|\mathcal{S}| - \eta}
    \nonumber \\
    \Rightarrow \eta &= \left(1 - \cfrac{\lambda}{\sqrt{\underline{\lambda}\overline{\lambda}}} \right) |\mathcal{S}|
    \label{eq:design_eta}
\end{align}
The above approximation is accurate when $\nu$ is sufficiently small: that is, the larger the ratio, $\zeta = \overline{\lambda} / \underline{\lambda}$, the more accurate it becomes.
On the other hand, if $\zeta$ is small, $\eta < 0$, violating its definition.
Therefore, it is necessary to design $(\underline{\lambda}, \overline{\lambda})$ such that $\zeta > (3 + \sqrt{5}) / 2$.
In the following experiments (and the previous paper's ones), $\zeta = 100$, so this condition is normally satisfied.
Note that although it is also possible to derive $\eta$ using numerical solvers without the approximation, that would complicate the implementation and likely reduce reproducibility due to numerical errors.

\subsection{Policy divergence}

Next, we select an alternative metric to Hellinger distance, which has a finite upper bound, in order to measure the discrepancy between policies.
The simplest answer is KL divergence; in fact, this can be considered to be the metric adopted in the simplified version with the assumption that each distribution is a Gaussian distribution with a fixed scale of one.
However, there is no guarantee to find analytical solutions for such divergences with the actual policy models.
The introduction of Monte Carlo approximation is therefore not ignorable, while it cannot guarantee non-negativity when KL divergence is approximated.
Although this is not a problem when optimizing a regularization scalarized by the mean operator and solved by a gradient method, non-negativity is necessary to theoretically handle Lipschitz continuity and to apply the scalarization in the next section.

Therefore, we adopt the extended KL divergence~\cite{nielsen2020non}, which guarantees non-negativity even when approximated by Monte Carlo method.
\begin{align}
    \mathrm{KL}^\prime(p(x) \mid q(x)) = \mathbb{E}_{p(x)}\left[ \ln \cfrac{p(x)}{q(x)} + \cfrac{q(x)}{p(x)} - 1 \right]
\end{align}
Here, $q(x)/p(x) - 1$ added into the expectation represents the equivalent transformation; this ensures that the inside of the expectation is non-negative, thereby guaranteeing non-negativity even when using Monte Carlo approximation.
Furthermore, this extension makes the inside minimum when $p(x) = q(x)$, bringing the proper regularization.

As a remark, since KL divergence is asymmetric, we expect the effectiveness of regularization to differ depending on the order in which $\pi(a \mid s)$ and $\pi(a \mid \tilde{s})$ are input~\cite{kobayashi2022optimistic}.
In this paper, we examine both patterns; however, considering the balance between smoothing and expressiveness, it might be preferable to input $\pi(a \mid s)$ on the right-hand side.
That is, $\pi(a \mid s)$ is strongly smoothed to represent the average behavior in the surrounding states, in accordance with the mass-covering property.
In addition, $\pi(a \mid \tilde{s})$ can follow the different local spaces, where the expressiveness is required, by limiting the smoothing due to the mode-seeking property.

\subsection{Scalarization}

\begin{table*}[tb]
    \caption{Comparison of implementations: \textcircled{1}--\textcircled{4} refer to the four components in the smoothing methods applied in the local space; they are used according to Alg.~\ref{alg:impl}.}
    \label{tab:impl}
    \centering
    \begin{tabular}{c|cccc}
        \hline\hline
        Method & \textcircled{1} Probe point $\tilde{s}$ & \textcircled{2} Weight $w$ & \textcircled{3} Loss $\ell_R$ & \textcircled{4} Scalarization
        \\
        \hline
        Original~\cite{kobayashi2022l2c2}
        & $s + \sigma \Delta s \epsilon$
        & $\lambda/(\|\epsilon\|_\infty + \nu)$
        & $(\sqrt{\pi(a \mid s)} - \sqrt{\pi(a \mid \tilde{s})})^2 / (2 \mathrm{sg}(\pi(a \mid \tilde{s})))$
        & Mean
        \\
        & $\epsilon \sim \mathcal{U}_{|\mathcal{S}|}(-1, 1)$
        & w/ eq.~\eqref{eq:design_lambda}
        & $a \sim \pi(a \mid \tilde{s})$
        &
        \\
        \hline
        Simple~\cite{huang2025learning,huang2026towards,zhao2026agile}
        & $s + \sigma \Delta s \epsilon$
        & $\lambda$
        & $(\mu_\pi(s) - \mu_\pi(\tilde{s}))^2$
        & Mean
        \\
        & $\epsilon \sim \mathcal{U}_{1}(-1, 1)$
        &
        &
        &
        \\
        \hline
        LCP~\cite{chen2025learning}
        & $\lim_{h \to 0} s + h$
        & $\lambda$
        & $((\ln \pi(a \mid \tilde{s}) - \ln \pi(a \mid s)) / h)^2$
        & Mean
        \\
        &
        &
        & $a \sim \pi(a \mid s)$
        &
        \\
        \hline
        Proposal
        & $s + \sigma \Delta s \epsilon$
        & $\lambda/(\|\epsilon\|_\infty^\eta + \nu)$
        & $\ln \pi(a \mid \tilde{s}) / \pi(a \mid s) + \pi(a \mid s) / \pi(a \mid \tilde{s}) - 1$
        & LogSumExp
        \\
        & $\epsilon \sim \mathcal{U}_{|\mathcal{S}|}(-1, 1)$
        & w/ eqs.~\eqref{eq:design_lambda} and~\eqref{eq:design_eta}
        & $a \sim \pi(a \mid \tilde{s})$ w/ reparameterization trick~\cite{kingma2014auto}
        &
        \\
        \hline\hline
    \end{tabular}
\end{table*}

Finally, we modify the method used to scalarize the regularization terms with respect to the batched data employed in calculating the main loss function.
While the conventional mean operation is simple and promotes overall smoothing by shifting the center of the loss distribution, it fails to adequately account for data at the tails of the distribution that should be smoothed more.
However, since this behavior is effective in ensuring expressiveness, it cannot be said that it is necessarily desirable to strictly suppress the worst-case data at the edges of the distribution to a certain level of smoothness.
Therefore, we introduce a smooth scalarization method that gives higher priority to cases where the regularization term is larger.

Specifically, a tilted loss or smooth Tchebycheff scalarization~\cite{li2023tilted,lin2024smooth} is employed.
This can be given as the following log-sum-exponential function.
\begin{align}
    \mathcal{L}_R = T \ln \sum_{i=1}^N \exp(w_i\ell_R(s_i, s_i^\prime) / T)
\end{align}
where, $w_i\ell_R(s_i, s_i^\prime) = \lambda \rho_\pi(s_i, \tilde{s})$ with $\tilde{s} \sim \mathcal{S}_{s_i,s_i^\prime}$, $N$ denotes the batch size, and $T > 0$ is the temperature parameter.
Although it is possible to leave room for adjustment by treating $T$ as a hyperparameter, this paper focuses on points where $w \ell_R \geq 0$, and introduce an implementation that treats the batch mean $\mu_R = \mathrm{sg}\left(N^{-1}\sum_{k=1}^N w_k \ell_R(s_k, s_k^\prime) \right)$ (where $\mathrm{sg}(\cdot)$ refers to a stop-gradient operation) as $T$.
This serves as a form of normalization and results in easy-to-interpret behavior: the regularization terms larger than $\mu_R$ are given higher priority.

In addition, since $w\ell_R$ can be decomposed into the weight $w$ and the policy divergence $\ell_R$, data points that are close to $s$ but are not yet sufficiently smoothed are given higher priority.
Conversely, data that is far from $s$ and already sufficiently smoothed, and thus requires little further smoothing, is implicitly excluded, allowing the model to regain the expressiveness at that point.
Note that this qualitative property holds because the policy divergence is designed to be non-negative: if negative values were allowed, the role of the weights would change depending on the sign (and also, $\mu_R$ cannot be utilized as $T$).

\subsection{Comparison and pseudo code}

When comparing the above modifications with the original implementation and others, the result is summarized in Table~\ref{tab:impl}.
The loss for LCP~\cite{chen2025learning} can be computed using automatic differentiation.
Note that only the loss for the proposed method uses the reparameterization trick~\cite{kingma2014auto} to sample $a$; this is because it addresses the asymmetry in KL divergence, but we do not consider it necessarily required.
Furthermore, the pseudocode for implementing them is described in Alg.~\ref{alg:impl}.
Note that, for the sake of simplicity, the optimization process for the value function has been omitted, although it can also be regularized by minimizing the squared errors with the same weights and the scalarization as the policy's ones, as in the original paper~\cite{kobayashi2022l2c2}.

\begin{algorithm}[tb]
    \caption{Pseudocode of policy optimization with the components in Table~\ref{tab:impl}}
    \label{alg:impl}
    \begin{algorithmic}[1]
        \State{Given batched data $\{(s_i, a_i, s_i^\prime, r_i)\}_{i=1}^N$}
        \State{Compute main loss $\mathcal{L}_\pi = N^{-1}\sum_{i=1}^N \ell_\pi(s_i, a_i, s_i^\prime, r_i)$}
        \For{$i=1,\ldots,N$}
            \State{Get $\tilde{s}_i$ by \textcircled{1}}
            \State{Compute $w_i$ with $\tilde{s}_i$ by \textcircled{2}}
            \State{Compute $\ell_{R,i}$ with $\tilde{s}_i$ by \textcircled{3}}
        \EndFor
        \State{Scalarize $w_i \ell_{R,i}$ ($i=1,\ldots,N$) as $\mathcal{L}_R$ by \textcircled{4}}
        \State{Update $\theta$ with the gradient of $\mathcal{L}_\pi + \mathcal{L}_R$}
    \end{algorithmic}
\end{algorithm}

\section{Simulations}

\subsection{Sensitivity analysis}

\begin{figure}[tb]
    \centering
    \includegraphics[keepaspectratio=true,width=\linewidth]{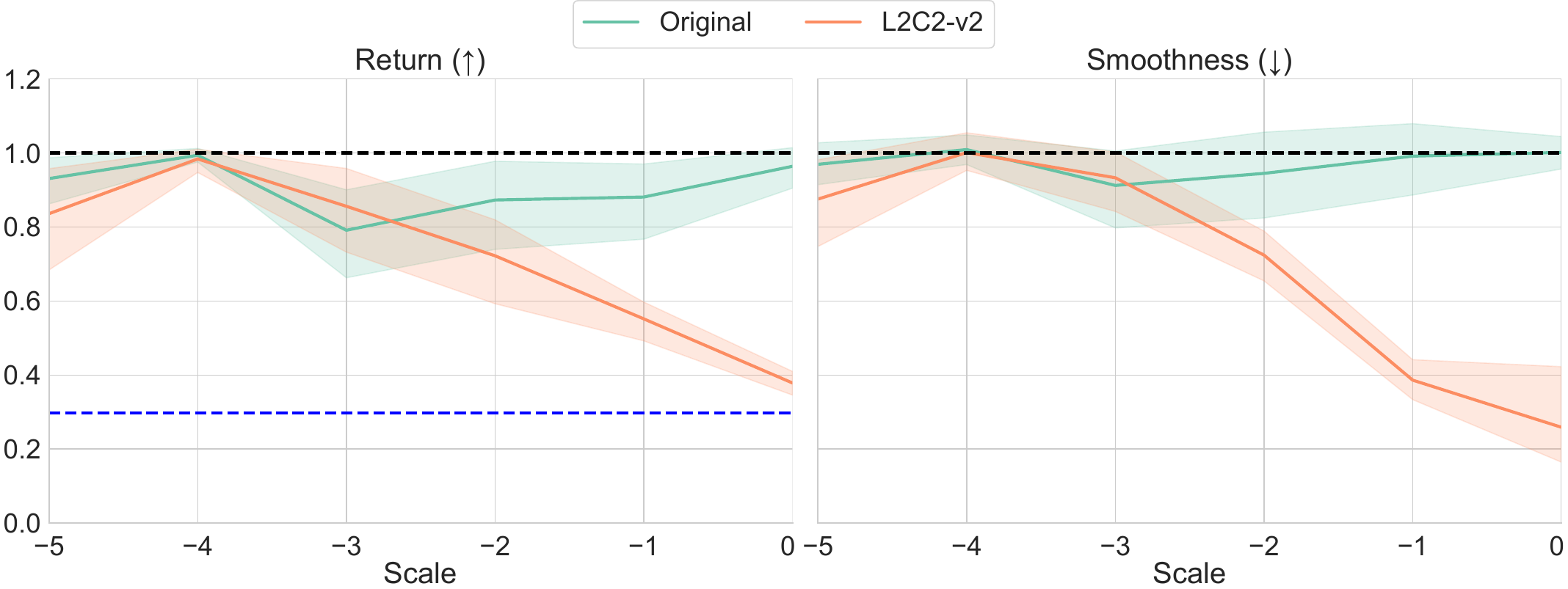}
    \caption{Sensitivity analysis of scale for regularization}
    \label{fig:sensitivity}
\end{figure}

The proposed L2C2-v2 has the same hyperparameters as the original L2C2.
Among these, the most influential one is expected to be the scale of $(\underline{\lambda}, \overline{\lambda})$.
To verify the extent of this influence, we solved \textit{Ant-v4} using the soft actor-critic (SAC)~\cite{haarnoja2018soft} implemented in CleanRL~\cite{huang2022cleanrl}; the results with 20 random seeds are shown in Fig.~\ref{fig:sensitivity}.
The left shows return and the right is smoothness (the episodic mean of $\|a_{t-1} - a_{t}\|_2^2$), while both horizontal axes are with the scale $c$ for $(\underline{\lambda}, \overline{\lambda})=(10^{c-2}, 10^{c})$.
Note that their scores are normalized by the case without regularization.
The blue dashed line indicates the return when the agent is capable of minimal walking, showing that walking itself has been acquired under all conditions.

Interestingly, the original L2C2 was not greatly affected by the scale and resulted in only a slight regularizing effect around $10^{-3}--10^{-2}$.
This is probably because, as noted above, the three factors mentioned prevent the gradients from effectively smoothing the policy.
On the other hand, as expected, L2C2-v2 promotes smoother motion as the scale increases.
However, since the goal of the Ant task is to maximize movement speed, the smoothing deteriorates the control performance.
Resolving this tradeoff would likely require designing $\sigma$.
However, since this depends heavily on task and situation, requiring automatic optimization, we will address this as future work.

\subsection{Comparison}

\begin{figure*}[tb]
    \centering
    \includegraphics[keepaspectratio=true,width=\linewidth]{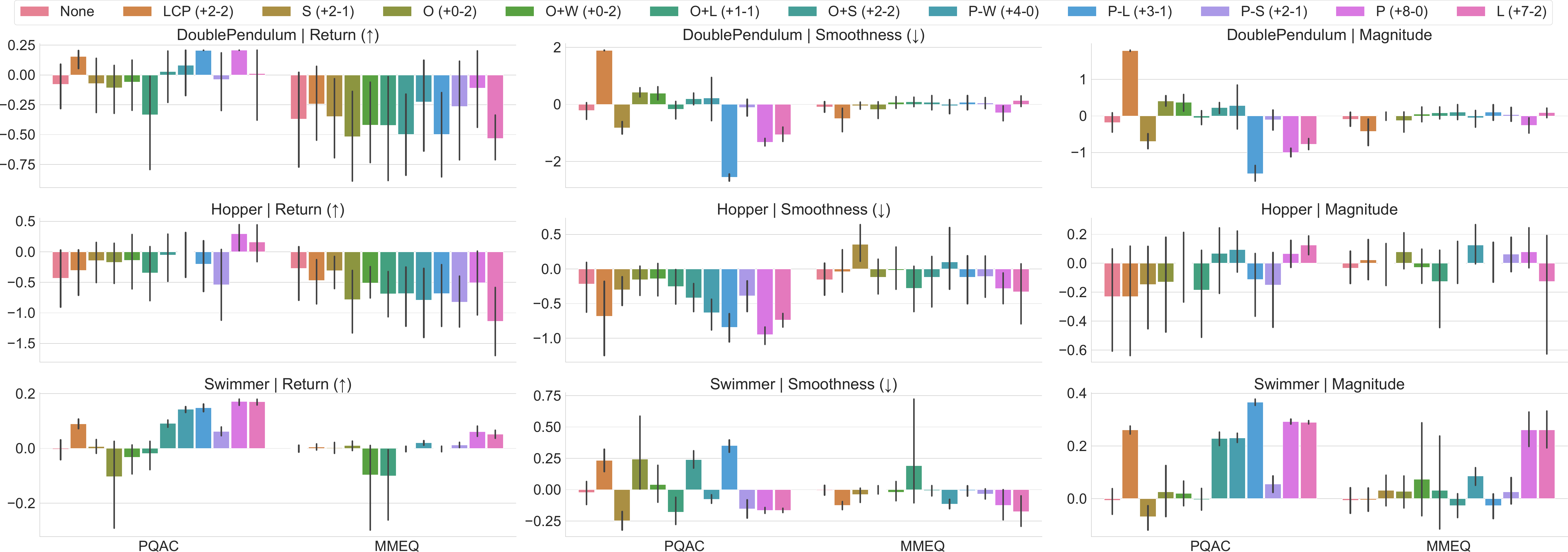}
    \caption{Test scores normalized by the ones at the None condition}
    \label{fig:summary}
\end{figure*}

\subsubsection{Setup}

Let us verify the impact of the proposed modifications on the control performance and the policy smoothness.
Three benchmark tasks are chosen from Gymnasium~\cite{towers2024gymnasium}: \textit{InvertedDoublePendulum-v5} (DoublePendulum), \textit{Hopper-v5} (Hopper), and \textit{Swimmer-v5} (Swimmer).
Note that Hopper increases \texttt{ctrl\_cost\_weight} to $0.5$ to promote smoothing at the reward level and to investigate the effects of interference; and Swimmer increases the number of segments to 15 (therefore, $|\mathcal{A}|=14$) to confirm the smoothing for multi-dimensional actions.
For each task, training is completed after 2,000 episodes.
Afterwards, 100 episodes are tested with the optimized policy and observation noise (i.e. Gaussian noise with a standard deviation of $10^{-2}$).
Three metrics, namely return, smoothness, and action magnitude (i.e. $\|a\|_2^2$), are recorded for every random seeds.

Two RL algorithms are employed as baselines: PQAC~\cite{kobayashi2026pseudo} with weighted log-likelihood, and the algorithm implemented in the literature~\cite{kobayashi2026flexible} (which we refer to in this paper as max-min entropy and Q-value (MMEQ)) based on SAC with reparameterization trick.
On top of these, we implement and compare the following six conditions:
\begin{itemize}
    \item \textbf{None}: Baseline without any regularization
    \item \textbf{LCP}: Regularization by LCP~\cite{chen2025learning} with $\lambda = 10^{-5}$
    \item \textbf{S}: Simplified version of L2C2 with $\lambda = 10^{-3}$ and $\sigma=1$
    \item \textbf{O}: Original implementation of L2C2
    \item \textbf{P}: The proposed method
    \item \textbf{L}: Modified by swapping $\pi(a \mid s)$ and $\pi(a \mid \tilde{s})$
\end{itemize}
In addition to them, to investigate the effects of each component, we also test cases where only one of the following is added or removed, totaling 12 conditions: \textcircled{2} Weight (as \textbf{W}), \textcircled{3} Loss (as \textbf{L}), and \textcircled{4} Scalarization (as \textbf{S}).

\textbf{O}, \textbf{P}, and \textbf{L} use common hyperparameters, with $(\underline{\lambda}, \overline{\lambda}) = (10^{-4}, 10^{-2})$ and $\sigma=1$.
Note that $\lambda$ of LCP was adjustd to ensure that training proceeds without issues in DoublePendulum, and the resulting scale change from its original paper was applied to other conditions as well to ensure fairness.
This scale change would be due to the degree of normalization within the originally-adopted RL algorithm and the model structure of the policy.

\subsubsection{Results}

The results with 20 random seeds are summarized in Fig.~\ref{fig:summary}.
Note that the results have been normalized and transformed logarithmically.
The numbers next to the labels indicate the count of positive or negative significant difference observed when each metric of each condition was compared to None using the Mann-Whitney test.

First, because LCP focuses on smoothing within the extremely small local space, it caused jitter in DoublePendulum with PQAC due to observation noise, and it did not necessarily achieve smoothing in other cases either.
The original L2C2 derived almost no benefit from smoothing, as confirmed in the sensitivity analysis.
Note that the excellent regularization results reported in previous studies~\cite{kobayashi2022l2c2,christmann2024benchmarking} are thought to be indirect effects resulting from the smoothing of the value function.
The simplified implementation achieved a certain degree of smoothing, with the exception of the Hopper with MMEQ.
Performance was almost maintained, and there was little change in the magnitude, so this approach might be useful for fine-tuning.

In contrast, the proposed method achieved greater smoothing in all cases and also increased performance on all tasks except for Hopper with MMEQ.
Furthermore, we observed several cases where this smoothing was achieved even as the magnitude increased, suggesting that it encouraged the emergence of behavior that differs from the baseline.
A similar trend can be observed in the swapped version, which is for evaluating the asymmetry of KL divergence; however, it did not match the performance of the proposed method probably due to the asymmetric regularization effect (i.e. the mass-covering or mode-seeking property).

Focusing on the number of significant differences, we can see that adding one component to the original method does not result in a significant improvement, whereas removing one component from the proposed method leads to a marked deterioration.
Furthermore, adjusting the weights had little effect, while replacing the scalarization was more effective.
Modifying the loss function is also critical because P-L in the Swimmer with PQAC failed smoothing most significantly.

\begin{figure}[tb]
    \centering
    \includegraphics[keepaspectratio=true,width=0.9\linewidth]{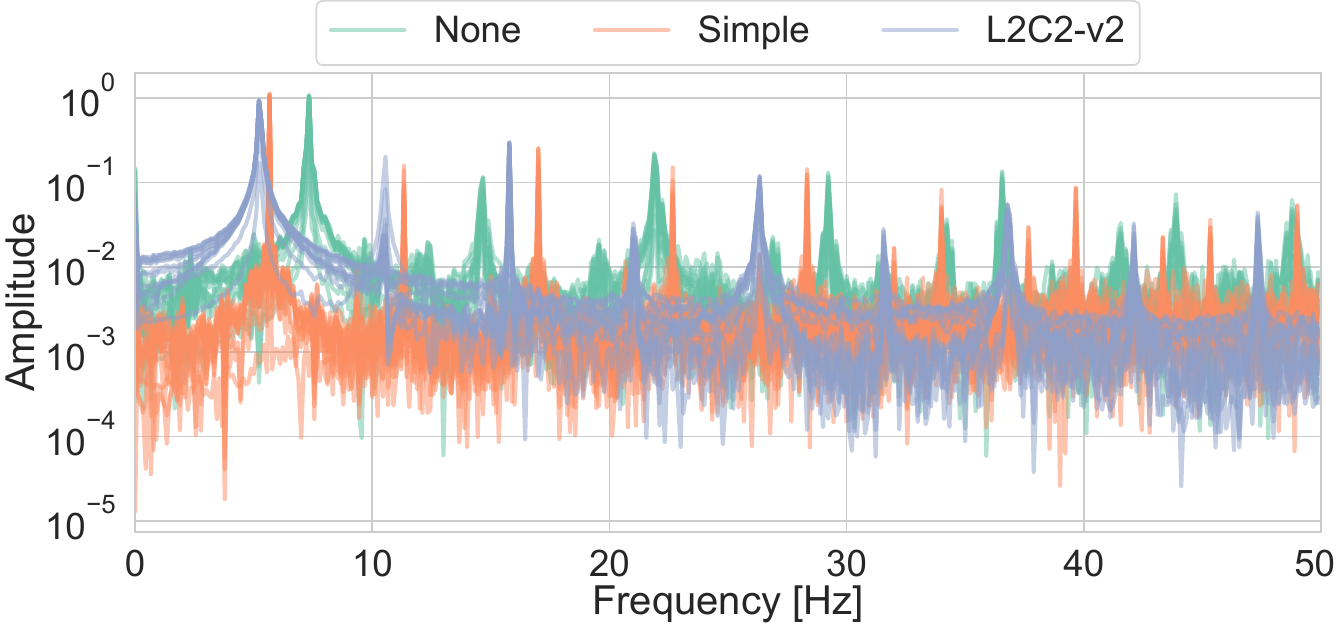}
    \caption{Frequency analysis for Swimmer with PQAC}
    \label{fig:frequency}
\end{figure}

To verify the differences in the emerging movements, a frequency analysis is performed on the action sequences of the Swimmer with PQAC, which exhibited the most distinctive trends (see Fig.~\ref{fig:frequency}).
For clarity, we have plotted only the non-regularized case (None), the simplified version (Simple) that was successfully smoothed, and the proposed method (L2C2-v2), all of which have similar return around 200.
First, it is clear that None exhibited a generally higher frequency, while the fundamental frequencies of Simple and L2C2-v2 were close, although such a slight difference widened as in higher harmonics.
L2C2-v2 tended to exhibit faster amplitude decay, whereas Simple showed decay similar to that of None.
Thus, L2C2-v2 acquired the motion dominated by the low-frequencies (i.e. coordination of all joints).

\section{Real-robot experiments}

\subsection{Setup}

Finally, to demonstrate the importance of smooth motion in the real world, we perform sim-to-real RL~\cite{rudin2022learning} on Unitree Go2, a quadrupedal robot, and evaluate its walking performance.
We use the Unitree's official repository, \texttt{unitree\_rl\_lab}, and train for a total of 20,000 iterations using the default settings (i.e. tracking control to randomly generated target twist commands, while retaining the reward terms related to motion smoothing).
This training uses \texttt{rsl\_rl}~\cite{schwarke2025rsl}, which has been slightly modified to calculate the loss provided by the proposed L2C2-v2 (and the simplified version).
In this experiment, we set $(\underline{\lambda}, \overline{\lambda})$ to $1/10$ of the values above to prioritize learning efficiency.

The optimized policy is deployed to the real robot, which then sequentially executes a total of six velocity commands: a 1.5-sec command to a single axis followed by a 0.5-sec pause, repeated for both positive and negative directions on all axes (x, y, and yaw).
The observations from 10 trials of this profile are statistically evaluated.
Specifically, the 10 step-wise metrics are averaged for each trial and the maxima in all the trials are used for normalization, then the obtained metrics are compared among None, Simple, and L2C2-v2.
Note that when calculating the first metric (i.e. tracking error) it is normalized by the magnitude of the maximum speed and the target speed in each axis, ensuring that errors at low speeds are not overlooked.
In addition, the final metric (i.e. work) is estimated from the product of the velocity and torque at each joint for evaluating energy efficiency.

\subsection{Results}

\begin{figure}[tb]
    \centering
    \includegraphics[keepaspectratio=true,width=\linewidth]{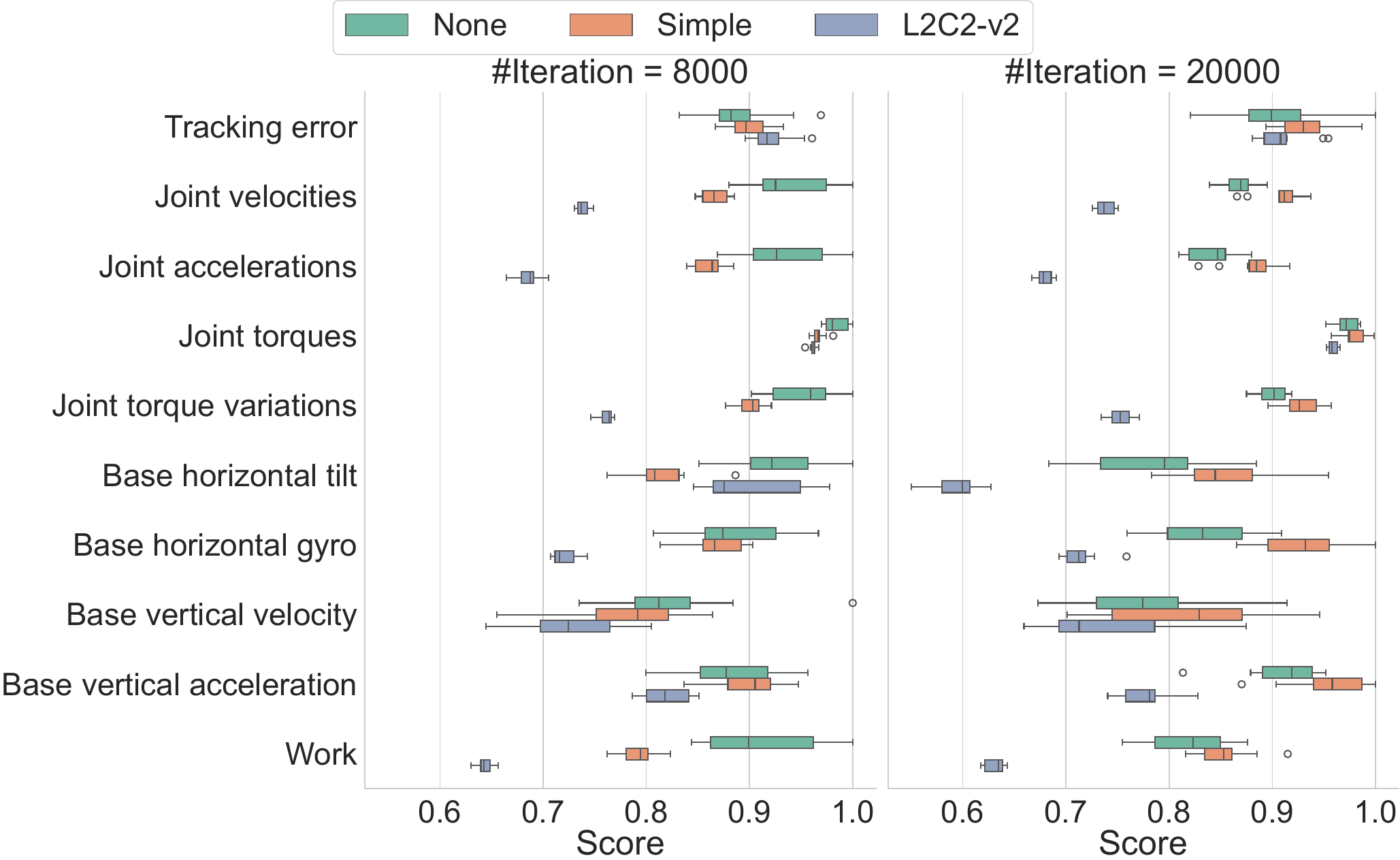}
    \caption{Results of 10 trials with the fixed velocity profile}
    \label{fig:sim2real}
\end{figure}

The results are depicted in Fig.~\ref{fig:sim2real}.
Note that to illustrate the progress of the training, we performed the same evaluation at 8,000 iterations and have included the results on the left.

During the first half of training, L2C2-v2 exhibited a larger tracking error than the others possibly due to a delay in learning, but eventually converged to a level not
significantly different to None.
This is because smoothing improved tracking accuracy for low-speed commands.
Indeed, the next four metrics regarding joint states represent the smoothness of motion, and all showed significant improvements with L2C2-v2 from the first half.
In particular, the fact that comparable tracking was achieved while suppressing joint velocities and torques indicates that more efficient walking has emerged, as evaluated in the final metric, namely work.

The remaining four metrics regarding the base link are related to walking stability, and again, significant improvements were confirmed in all of them finally, although the base remained tilted during the first half.
In particular, the reduction in vertical acceleration indicates that the impact force at landing feet was reduced while the others increased eventually.
In other words, L2C2-v2 can mitigate one of the sim-to-real gaps, in addition to jitter.

As a remark, although Simple maintained a tracking error comparable to that of None during the first half while benefiting from a slight amount of smoothing, its scores eventually deteriorated across the board.
In contrast, None and L2C2-v2 generally achieved better final scores, demonstrating a refinement in their motions.
This suggests that Simple's weak regularization destabilized the training process.
The reason L2C2-v2 did not experience instability is also implied that its stronger regularization acted as a constraint, allowing the exploration to proceed under the constraint.

The benefits of this walking stabilization become particularly evident when more extreme velocity profiles are applied.
The attached video\footnote{\url{https://youtu.be/TyFuVzj6QLw}} demonstrates what happens when we remove the pauses between velocity commands and double the lateral velocity.
In the vanilla implementation, the robot was unable to maintain balance and fell over when subjected to sudden lateral velocity commands due to a small lateral margin of stability.
In contrast, by achieving smoother motion with the proposed L2C2-v2, the robot was able to stably follow such an extreme velocity profile.
While strategies involving random large velocity commands to escape from a stuck state have proven effective in competitions when combined with a higher-level navigation module~\cite{irie2025rough}, robustness against such sudden changes in velocity commands is expected to broaden the range of possible strategies.

\section{Conclusion}

This paper identified three issues hidden in the original implementation of L2C2, a regularization method for policy smoothing, and provided theoretically sound solutions to them.
As a result, the proposed method, so-called L2C2-v2, achieved effective smoothing across multiple benchmarks and RL algorithms while also confirming improvements in control performance.
Furthermore, when applied to sim-to-real RL to investigate the response to sudden changes in target velocity for quadrupedal walking, the motion became noticeably smoother and more energy-efficient, and stabilization was achieved as a secondary benefit.
In fact, even with the extreme velocity profile that causes a fall with the vanilla policy, L2C2-v2 made tracking successfully stable.

The effectiveness of regularization depends heavily on the design of hyperparameters, as shown in the tradeoff between control performance and smoothness, which has been found by the sensitivity analysis.
We therefore plan to investigate mechanisms for automatically tuning those hyperparameters in L2C2-v2 in the near future.
Furthermore, considering that the emerging motion can diverge depending on the degree of smoothing, we believe that by training a certain type of multi-objective policy~\cite{alegre2025amor} that incorporates these hyperparameters as conditions, the robot will be able to elicit motion with an appropriate degree of smoothness at inference.
Based on these improvements, we aim to be able to smoothly accomplish more dynamic and complex robotic tasks.

%
%
%
\bibliographystyle{IEEEtran}
{
\bibliography{biblio}
}


\end{document}